\documentclass[letterpaper, 10pt, conference]{ieeeconf}  
\IEEEoverridecommandlockouts                              
\usepackage[pdftex]{graphicx}
\usepackage{amsmath}
\usepackage{amsfonts}
\usepackage{amssymb}
\usepackage[lined, ruled,linesnumbered]{algorithm2e}
\usepackage{subfigure}
\usepackage{url}
\usepackage{algorithm2e}
\usepackage{pifont}
\usepackage{color}
\usepackage{tabularx}
\usepackage[table]{xcolor}
\usepackage{pgfplots}

\long\def\/*#1*/{}

\newcolumntype{A}{>{\centering\arraybackslash}X}

\makeatletter
\def\blx@maxline{77}
\makeatother

\title{\LARGE \bf
Underwater Image Haze Removal and Color Correction with an \\ Underwater-ready Dark Channel Prior}

\author{
	\authorblockN{Tomasz {\L}uczy{\'n}ski and Andreas Birk}
	\authorblockA{Robotics Group \\
		Computer Science \& Electrical Engineering,\\
		Jacobs University Bremen, Germany\\
		t.luczynski@jacobs-university.de}
}

\begin{document}

\maketitle
\thispagestyle{empty}
\pagestyle{empty}

\begin{abstract}
	Underwater images suffer from extremely unfavourable conditions. Light is heavily attenuated and scattered. Attenuation creates change in hue, scattering causes so called veiling light. General state of the art methods for enhancing image quality are either unreliable or cannot be easily used in underwater operations. On the other hand there is a well known method for haze removal in air, called Dark Channel Prior. Even though there are known adaptations of this method to underwater applications, they do not always work correctly. 
	This work elaborates and improves upon the initial concept presented in \cite{oceansAnchorageDCP}. A modification to the Dark Channel Prior is proposed that allows for an easy application to underwater images. It is also shown that our method outperforms competing solutions based on the Dark Channel Prior. Experiments on real-life data collected within the DexROV project are also presented, showing robustness and high performance of the proposed algorithm.
\end{abstract}

\section{Introduction and related work} 
\label{sec:introduction}
Cameras are very popular sensors in underwater applications. They are relatively cheap, provide data with high update rate and may be used for tasks ranging from general exploration to object recognition and manipulation. On the other hand underwater imaging suffers from exceptionally bad light conditions and significant refraction based distortions. Image deformations may be corrected with calibration, using for example the recent Pinax model \cite{OE_Luczynski_2017_Pinax_mono}. However the problems related to underwater light propagation are much more complex. Light is strongly attenuated and scattered. An underwater image formation model is described in \cite{Jaffe1990}. Further analysis of the image formation process can be found in multiple subsequent papers, e.g.,\cite{SchechnerKarpel2005,ChiangChen2012,YamashitaFujiiKaneko2007}. This classic image formation model model was recently validated and some improvements were suggested in \cite{AkkaynakCVPR2017,AkkaynakCVPR2018}. This is discussed in details in Section \ref{sec:image_formation}.

\begin{figure}[htbp]
	\centering
	\includegraphics[width=\linewidth]{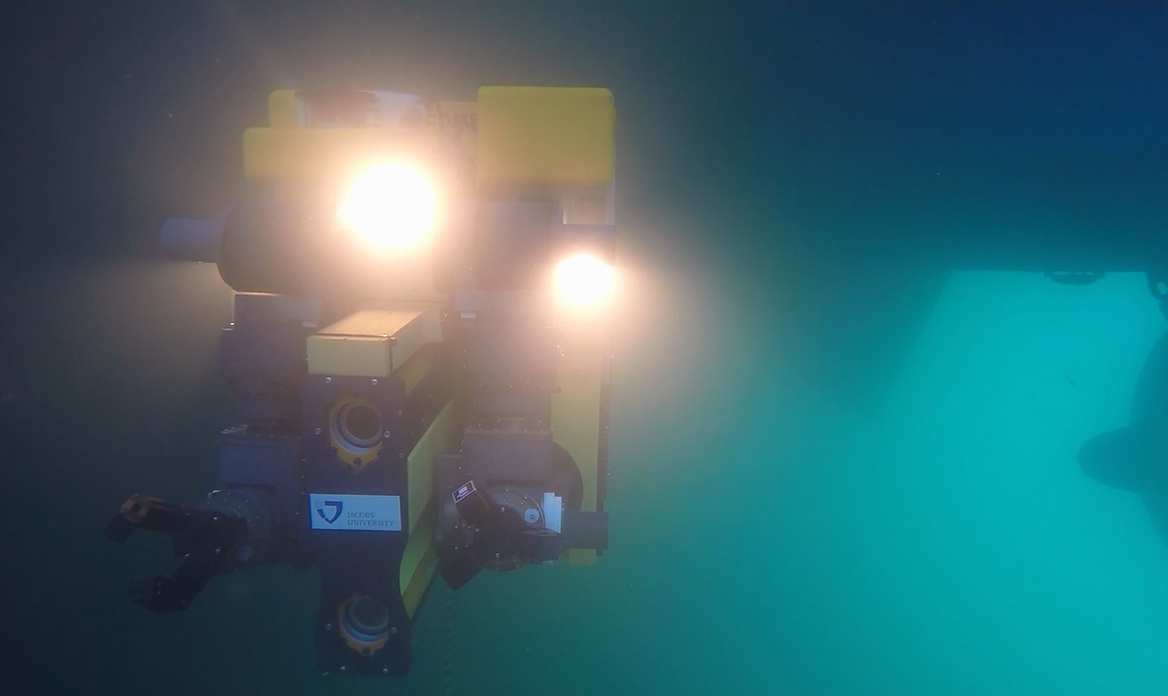}
	\caption{The DexROV project vehicle. Note how the lights generate significant amount of haze in front of the cameras.}
	\label{fig:dexrovpromo}
\end{figure}

Many attempts have been made to improve the quality of underwater images. An overview of these results is given for example in \cite{Schettini2010}. Naively one could think that white balance algorithms could be used to improve colours in the underwater images. Unfortunately this approach usually fails, as the change in colours is depth dependent and therefore the correction would depend on the distance to the observed object, which may be different for different parts of the image. Specialized white balance algorithm for underwater conditions was proposed in \cite{BiancoNeumannUW_WB}. Here the white balance is performed in the Rudermann colour space. Furthermore the new colours are calculated taking into account only neighbourhood of a given pixel, not the full image. This white balance algorithm is applied only to a patch where all the points are approximately equidistant to the camera. Usually this method gives good results with the exception of parts of the image, where there is a rapid change in distance, e.g. on the edges of some underwater structure with open water in the background.
As identified in \cite{SchechnerKarpel2005} and \cite{TreibitzSchechner2009} backscattering is the dominant degradation component. A physically accurate, polarization based method for haze removal is proposed in \cite{SchechnerKarpel2005}. However this method requires artificial light with a polarizer and to take two images from the same viewpoint with the analyser being physically differently oriented. Therefore this method is difficult, or in some cases impossible, to apply, especially one moving vehicles. On the other hand there is a well known method for haze removal from images taken in air, called Dark Channel Prior (DCP) \cite{HeDCP}. Some attempts to adapt this method for underwater conditions were made \cite{UDCP, RDCP}. Details of this research is discussed later.
Work presented in this paper is motivated by the DexROV project (``Dexterous ROV: effective dexterous ROV operations in presence of communication latencies'') funded within EU Horizon2020 programme. In this project, an ROV is supposed to perform multiple surveillance and manipulation tasks in semi-autonomous mode at significant depths (with no natural light available). When using artificial lightning backscattering effects may be especially visible (see Fig. \ref{fig:dexrovpromo}). On the other hand, 3D stereo reconstruction is to be used, therefore haze removal, even at a cost of degrading the color information, is crucial. As initial tests showed, state of the art haze removal methods did not lead to sufficient results for robust stereo processing - they either cannot be applied to a moving ROV (\cite{SchechnerKarpel2005,SchechnerAverbuch2007}) or the improvement was considered to be insufficient (\cite{UDCP, RDCP}).
The main contributions of this paper are:
\begin{itemize}\itemsep0pt
	\item identifying the reasons, for which the Dark Channel Prior and its existing modifications do not perform as well as expected in the underwater environment,
	\item proposing a new adaptation to the Dark Channel Prior that allows for easy application underwater. This is tested on real data recorded in various waters proving robustness and good performance.
\end{itemize}
 
The rest of the paper is structured as follows. Section \ref{sec:image_formation} details the underwater image formation model and the influence of all the system and environmental factors. Section \ref{sec:DCP} shortly introduces Dark Channel Prior as our approach is based on this work \cite{HeDCP}. Existing modifications of this method are also shortly discussed. Finally Sections \ref{sec:underwaterDCP} present initial results from \cite{oceansAnchorageDCP} which were proof of concept underline the method presented here, discussed in details in Section \ref{sec:final_alg}. Finally, Section \ref{sec:final_results} presents the results of our method being applied to real life data collected in various conditions in Mediterranean and Adriatic seas. Section \ref{sec:conclusions} concludes our findings.

\section{Underwater Image Formation}
\label{sec:image_formation}

\begin{figure}[htbp]
	\centering
	\includegraphics[width=\linewidth]{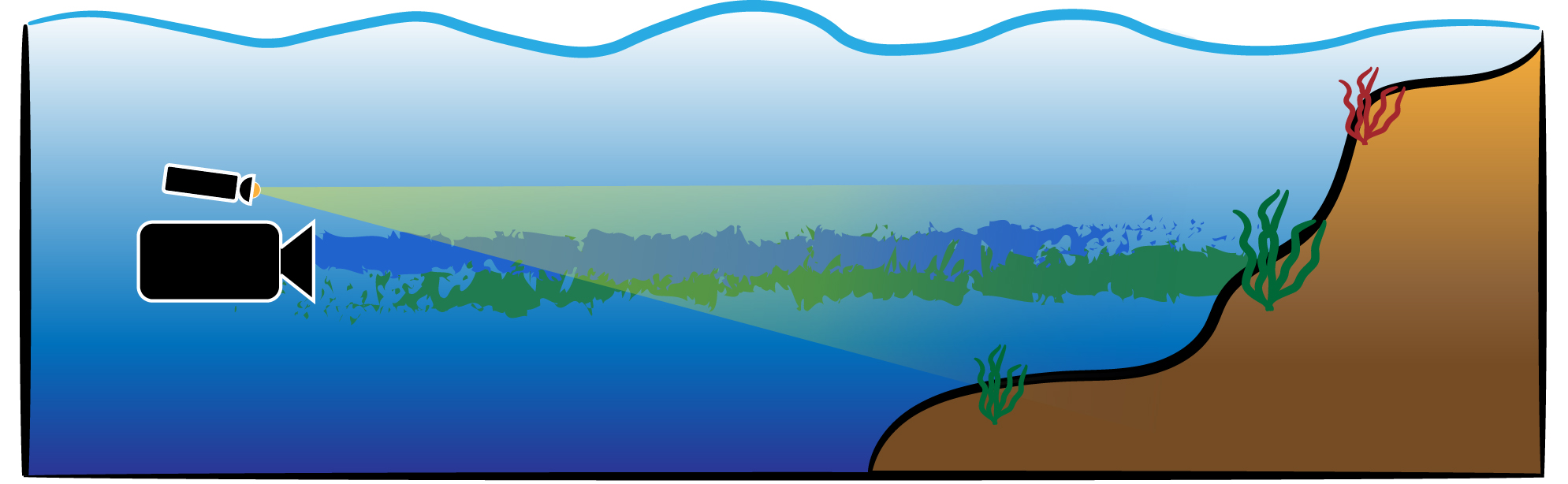}
	\caption{Underwater image formation scheme: green denotes light reflected from the object. This signal is getting weaker with distance, as the light is attenuated and scattered. Blue shows the veiling light caused by the artificial source and natural light coming from the surface. It is scattered and partially reflected back to the camera. The backscatter signal is getting stronger with the distance and may, eventually, dominate the direct transmission and occlude the observed scene. The goal of the haze removal is to remove the backscattered signal (blue), so that even if the remaining direct signal (green) is weak it is possible to retrieve the information. }
	\label{fig:image_formation}
\end{figure}
An underwater image formation model is described in multiple papers \cite{SchechnerKarpel2005,Jaffe1990}. For our work the following  model is used. The signal forming the image is a sum of three components:
\begin{align}
E_T=E_d+E_f+E_b
\end{align}
$E_T$ is a total radiance sensed by the camera. The first component forming $E_T$ is the direct transmission $E_d$ of the radiance reflected from the object $E_o$. Light, after being reflected from the scene, is attenuated. Attenuation is responsible for absorbing light with the distance from the source. This absorption depends on the wavelength, hence causing change in perceived colour of the observed scene:
\begin{align}
E_d=E_o e^{-c_{\lambda}r}
\end{align}
Where $r$ is a distance and $c$ is a total attenuation coefficient. Subscript denotes that $c$ depends on the wavelength. This equation also explains why white balance may only be used to correct registered colours if all registered points are approximately equidistant to the camera. In this case the shift in colour is similar for all the points and therefore white balance may succeed. However, it is important to remember, that applying white balance to the image will not restore true colours - it will make the image look more naturally, but it is based on some general assumption (e.g. so called grey world assumption) which may not be true for a given picture an therefore one should not expect to get an accurate colour reconstruction.
\begin{figure}[htbp]
	\centering
	\includegraphics[width=0.9\linewidth]{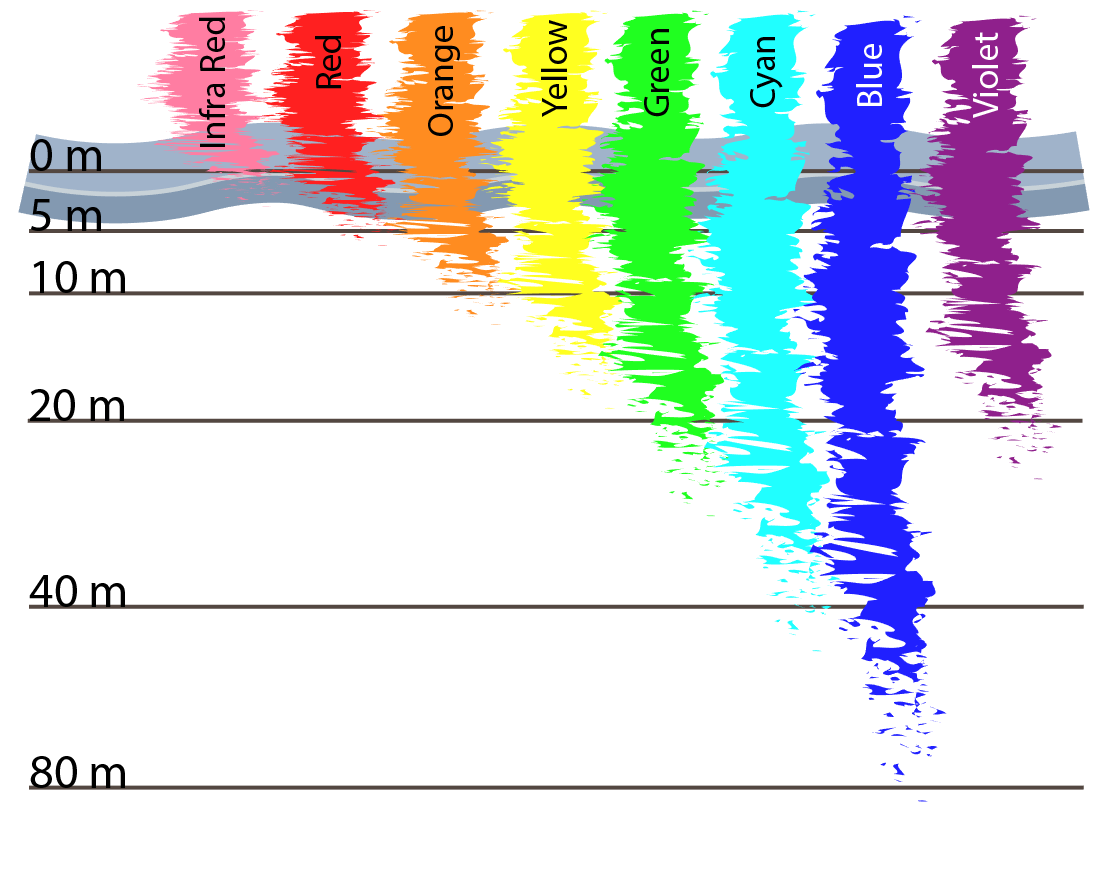}
	\caption{Attenuation of light underwater. }
	\label{fig:attenuation}
\end{figure}

The two remaining components of $E_T$ are caused by scattering and can be further divided into two phenomena: forward $E_f$ and back scattering $E_b$. The underlying physics for both is the same but the source of the scattered signal differs, they therefore have different influences on the image. Forward scattering occurs when light reflected from the scene scatters on its way to the camera. This results in slight blur in the image that can be described by convolution:
\begin{align}
E_f=E_d \ast g_r
\end{align}
Where $g_r$ is a point spread function (PSF) and is parametrized with distance $r$. There are several models of the underwater PSF \cite{Voss:91}, but their details do not matter to the work presented here and they will hence not be discussed any further.

Finally, backscattering, sometimes called the veiling light, occurs when an ambient light is being scattered and reflected back to the camera, adding a new signal component to the registered image. The further the scene from the camera, the stronger and more visible the backscattering component gets (compare Fig. \ref{fig:image_formation}). It can be compared to observing the environment in the fog. When no natural light is present the amount of backscattered signal depends on the overlap between the line of sight and the cone of artificial light being used. For that reason it may be desirable to use light separated from the camera to illuminate the scene from the side. However in most practical applications, e.g., when using ROV, the light is placed close to the camera, thus creating relatively strong backscatter. Backscatter caused by ambient light may be calculated as:
\begin{align}
E_b=B_{\inf}(1-e^{-c_{\lambda}r})
\end{align}
Where $B_{\inf}$ is a background signal, i.e., what would be seen if the camera was looking into open water with no objects in front of it. It is important to note, that it depends on light conditions, therefore may change, especially when only natural light is present. Furthermore, the bigger the distance of the camera to the observed scene, the stronger the veiling light. Because of that, the image cannot be corrected as a whole - for each pixel a different distance may apply, therefore different correction operations per pixel are necessary.

From the practical perspective, when 3D stereo reconstruction is the main goal, attenuation, forward and back scattering have different influence on the sensed image. Attenuation changes the color and lowers the overall intensity but does not influence the stereo 3D reconstruction significantly. Forward scattering introduces a slight blur, effectively smoothing the image, which may be beneficial for matching correspondences between the images. Finally back scattering was identified \cite{SchechnerKarpel2005} as the main source of the image degradation and reducing this phenomenon is therefore most important for accurate 3D reconstruction.

In the recent work \cite{AkkaynakCVPR2017,AkkaynakCVPR2018} the image formation model presented here was reviewed and improved. The numerical space within which the attenuation coefficients should be found was specified, based on the prior measurements of optical properties of different water bodies. Furthermore using wideband attenuation coefficients - for red, green and blue channels - was identified to be insufficient. Similar analysis was performed for backscattering signal. This research influences the image formation model significantly, but is irrelevant to this work, as we do not try to reconstruct the attenuation coefficients.

\section{Dark Channel Prior}
\label{sec:DCP}

The mechanism causing veiling light underwater is also present in air, only on a much smaller scale. It is caused by water and dust particles and may be observed on larger distances. In recent years, the Dark Channel Prior (DCP) \cite{HeDCP} became a popular method in this context. The dark channel $I^{dark}$ of a given image $I$ is defined as:
\begin{align}
I^{dark}(x)=\min_{y \in \Omega(x)} \left( \min_{c \in \{ r,g,b \}} I^c(x) \right)
\end{align}
The dark channel is formed by taking a minimum value from all the color channels within a patch around the given pixel. The Dark Channel Prior is based on the observation that in most of the non-sky patches (in outdoor images) at least one color channel has some pixels with very low intensity. Therefore:
\begin{align}
I^{dark} \rightarrow 0
\end{align}
For the sake of a more compact notation the double $min$ operator used to calculate the dark channel will be further denoted as:
\begin{align}
\min_{y \in \Omega(x)} \left( \min_{c \in \{ r,g,b \}} ... \right)=DC(...)
\label{eq:dcpop}
\end{align}

As shown in \cite{HeDCP} bright regions in the dark channel on non-sky regions appear due to the backscattering. Therefore it can be used to estimate and correct its influence on the image. Of course this method has its limitations, e.g., it cannot be used to process a snowy scene or a view at big white wall from close distance. However, in air the need for correcting backscattering occurs only when imaging outdoor scenes at long ranges. Therefore, DCP can be successfully used in these cases.
The haze removal in (long range) outdoor scenes with DCP assumes a similar model as in underwater vision:
\begin{align}
J(x)= I(x)t(x)+A(1-t(x))
\label{eq:transmissionAir}
\end{align}
where $J$ is an image without the haze, $A$ is a global atmospheric light - which corresponds to $B_{\inf}$ - and $t(x)$ is a transmission function:
\begin{align}
t(x)=e^{-\beta_\lambda r}
\end{align}
$\beta$ corresponds to the underwater attenuation coefficient $c$. In the first step $A$ is estimated. It is based on an other assumption, namely that atmospheric light is white and is calculated by taking the most haze-opaque region of the image (which can be found as the brightest region of the dark channel image). Then both sides of (\ref{eq:transmissionAir}) are normalized with $A$:
\begin{align}
\frac{J(x)} {A}= t(x) \frac{I(x)} {A} + 1-t(x)
\label{eq:transmissionAirNorm}
\end{align}
Then, the dark channel is calculated on both sides:
\begin{align}
DC \left( \frac{J(x)} {A} \right)= t(x) DC \left( \frac{I(x)} {A} \right) + 1-t(x)
\label{eq:transmissionAirNormDCP}
\end{align}
Using the dark channel prior on the haze-free image in the equation:
\begin{align}
DC \left( \frac{I(x)} {A} \right) = 0
\end{align}
This finally leads to the estimation of the transmission function:
\begin{align}
t(x) =1-DC\left( \frac{J(x)} {A} \right)
\end{align}
In the last step, the estimation of the transmission function is refined with soft matting \cite{soft_matting}. This step improves the quality of the corrected image, but also takes quite a long time to compute making the online application of this method impossible.

\begin{figure}[htbp]
	\centering
	\includegraphics[width=\linewidth]{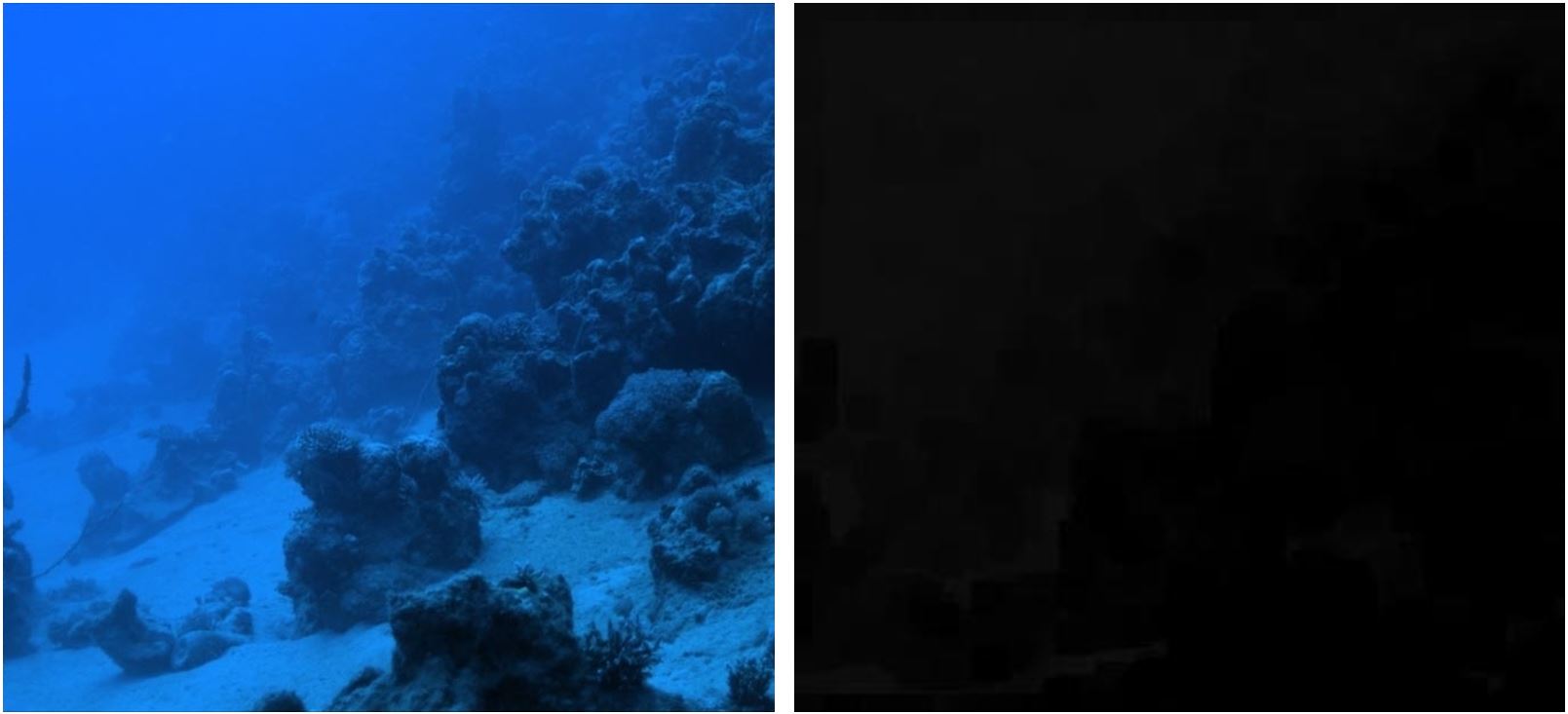}
	\caption{A raw underwater image (left) and its dark channel (right). Due to the high attenuation of the red light,  the dark channel is almost completely black. The image is taken from \cite{SchechnerKarpel2005,SchechnerAverbuch2007}; the image is shared for non-commercial use: http://webee.technion.ac.il/~yoav/research/underwater.html. It is one of the images used here for comparing our results to other DCP-based methods and the physically accurate approach presented in \cite{SchechnerKarpel2005,SchechnerAverbuch2007}. }
	\label{fig:origDC}
\end{figure}

As noticed earlier, the haze model underlying the dark channel prior correction method corresponds to the underwater backscatter model. Therefore it is immediately obvious that there is a significant potential in this method for underwater applications where the haze in the images is much more pronounced. Unfortunately, the dark channel prior cannot be used directly on underwater images. The method assumes that the backscatter component is white, which is true in air, but the heavy attenuation underwater causes the red wavelength to disappear very quickly (compare Fig. \ref{fig:attenuation}), leaving a (distance dependent) blueish hue of the veiling light. This also causes the dark channel to be completely black. If brighter regions appear, they usually correspond to the light patches of, e.g., sediment close to the camera (Fig.\ref{fig:origDC}).

\begin{figure}[htbp]
	\centering
	\includegraphics[width=\linewidth]{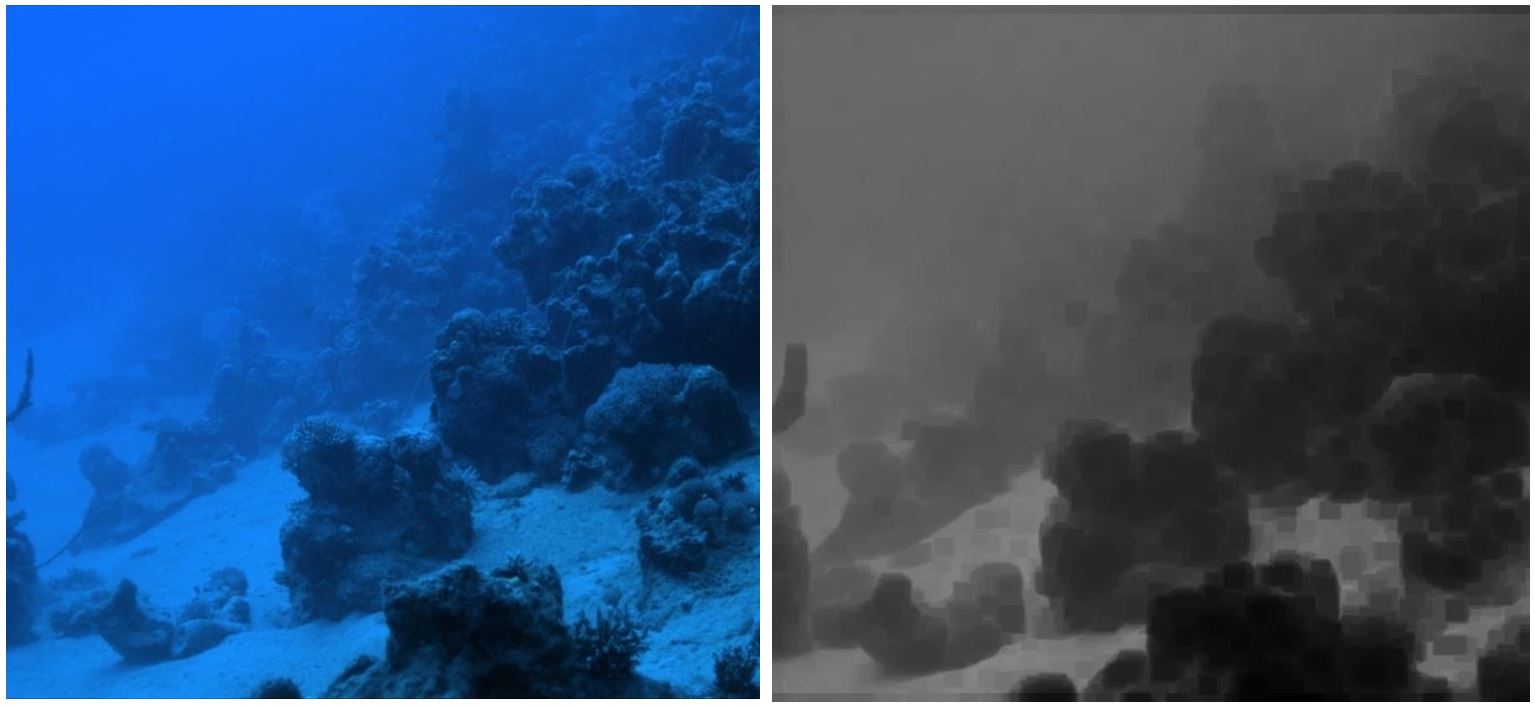}
	\caption{Raw underwater image (left) and its underwater dark channel calculated with \cite{UDCP} (right). There is a visible improvement compared to the classic DCP, but the result is still not ideal - the heavily hazed part of the image is not recognized as such. }
	\label{fig:UDC}
\end{figure}

Two major attempts to adjust the dark channel prior to underwater conditions were made before. In \cite{UDCP} it was noticed that attenuation of the red light is very strong. Therefore it was proposed to use green and blue channels only. The authors called this method Underwater Dark Channel Prior (UDCP). This modifies (\ref{eq:dcpop}):
\begin{align}
\min_{y \in \Omega(x)} \left( \min_{c \in \{ g,b \}} ... \right)=UDC(...)
\label{eq:udcpop}
\end{align}
Results, when using UDCP, are a bit better than those with DCP but the calculated dark channel is still not correct (Fig. \ref{fig:UDC}). The brightest region in the dark channel is the patch of sand close to the camera. This is understandable, as DCP in general does not handle regions like that well. A more severe drawback is that the heavily hazed background of the image is relatively dark, where in the optimal case it should be white.

\begin{figure}[htbp]
	\centering
	\includegraphics[width=\linewidth]{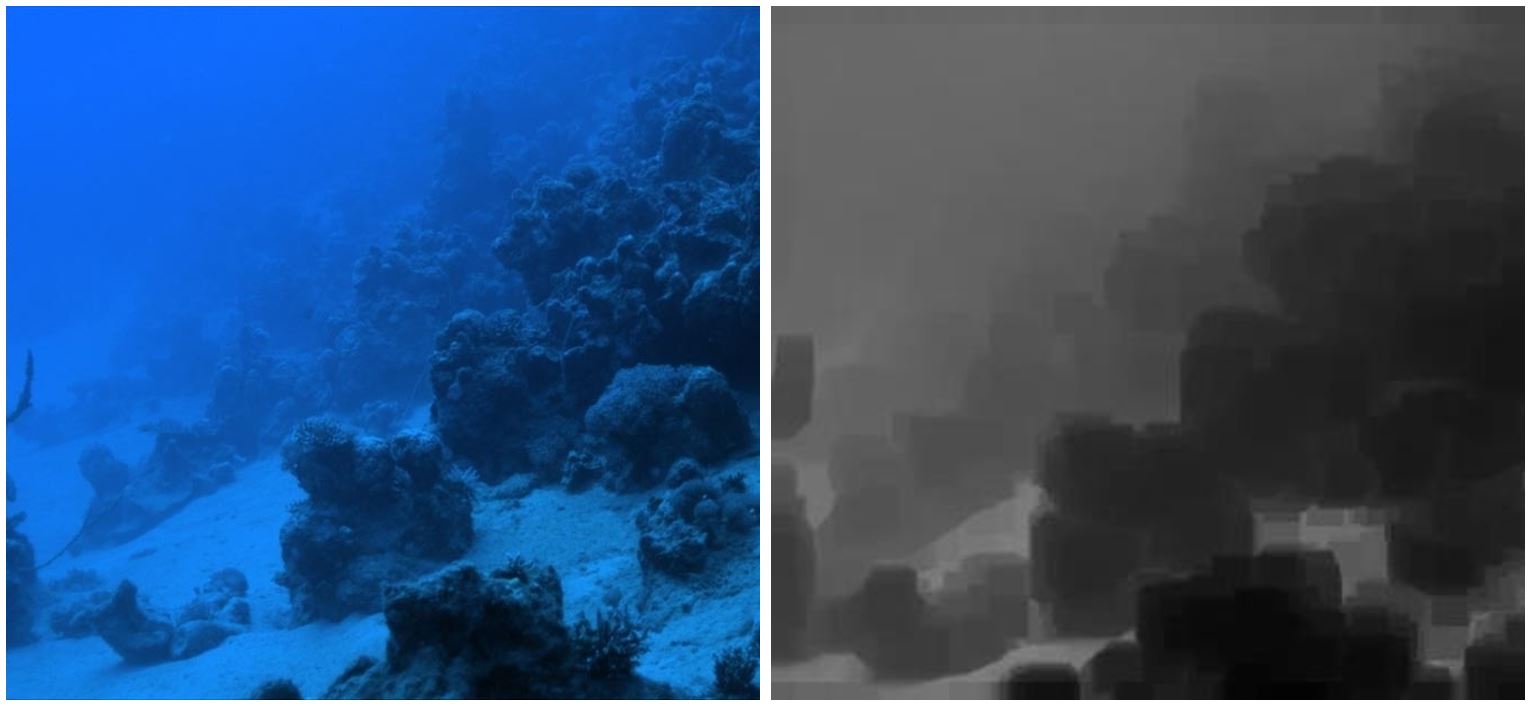}
	\caption{Raw underwater image (left) and its underwater dark channel calculated with \cite{RDCP} (right). There is a visible improvement comparing to the classic DCP, but the result is similar to the UDCP method and still not correct - the heavily hazed part of the image on is not recognized as such. }
	\label{fig:RDC}
\end{figure}

The authors of \cite{RDCP} build on the observation that the red light is attenuated so strongly that its presence implies a weak backscattering signal. Therefore, the complement of the red channel is used when calculating the dark channel. This method is here referred to as Red-Dark Channel Prior (RDCP). Equation (\ref{eq:dcpop}) is therefore modified to:
\begin{align}
\min_{y \in \Omega(x)} \left( \min_{c \in \{ 1-r,g,b \}} ... \right)=RDC(...)
\label{eq:udcpop}
\end{align}
Similar to UDCP, this method shows some improvements over DCP, but the dark channel calculated with this method is far from optimal (Fig. \ref{fig:RDC}).

\begin{figure*}[htbp]
	\centering
	\includegraphics[width=\linewidth]{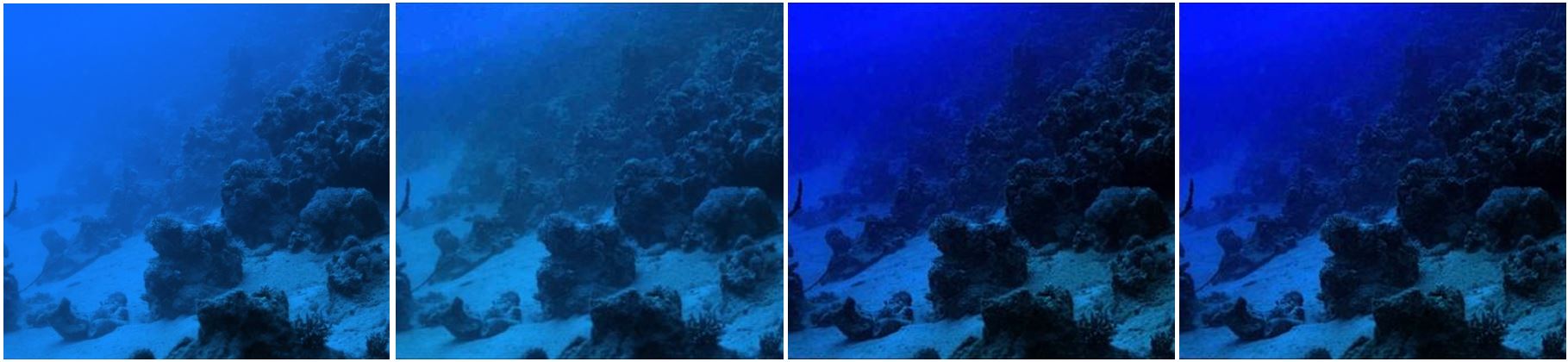}
	\caption{Comparison of the literature dark channel methods. From the left: raw image, classic DCP \cite{HeDCP}, UDCP \cite{UDCP} and RDCP \cite{RDCP}}
	\label{fig:SoAcomparis}
\end{figure*}

For example results of haze removal for all three methods please refer to Fig. \ref{fig:SoAcomparis}.

\section{Initial experiments}
\label{sec:underwaterDCP}
This section presents the initial experiments and discussion of the problem, as presented in \cite{oceansAnchorageDCP}.
\subsection{New approach}
As much as we agree with the observations made by authors of \cite{UDCP} and \cite{RDCP}, we believe that this problem should be tackled from a different angle. The observation underlying the original DCP can be formulated as follows: at least one color channel has some pixels with very low intensity for most of the non-sky patches. This can be reformulated to: the stronger the backscattering component, the whiter the region gets. Formulating the problem like this gives a better intuition for the adjustment of DCP to underwater: the backscattered light is (predominantly) blue in the underwater scenario, not white. Furthermore, it is only important for the proper estimation of the global atmospheric light $A$. If that step is performed correctly, the normalization step (\ref{eq:transmissionAirNorm}) removes the influence of the color light. 

\begin{figure}[htbp]
	\centering
	\includegraphics[width=0.7\linewidth]{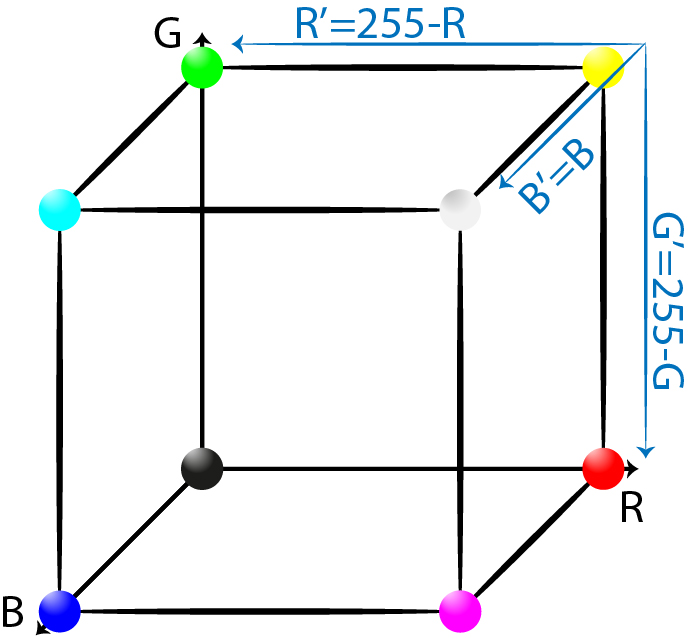}
	\caption{Proposed color transformation for easy and accurate estimation of $B_{\inf}$.}
	\label{fig:rgbcube}
\end{figure}

Therefore, the core of DCP does not need to be modified, but the method of estimating $A$ needs to be changed instead. Before estimating the global atmospheric light, all the colors in the image should be shifted in the color space so that blue becomes white. After this modification $A$ may be estimated from the dark channel of the modified image and the rest of the procedure remains as in \cite{HeDCP}. The modified procedure is as follows. To transform blue into white, the RGB coordinate system is shifted (compare Fig. \ref{fig:rgbcube}):

\begin{align}
R'=255-R \\
G'=255-G \\
B'=B\ \ \ \      
\end{align}

\begin{figure}[htbp]
	\centering
	\includegraphics[width=\linewidth]{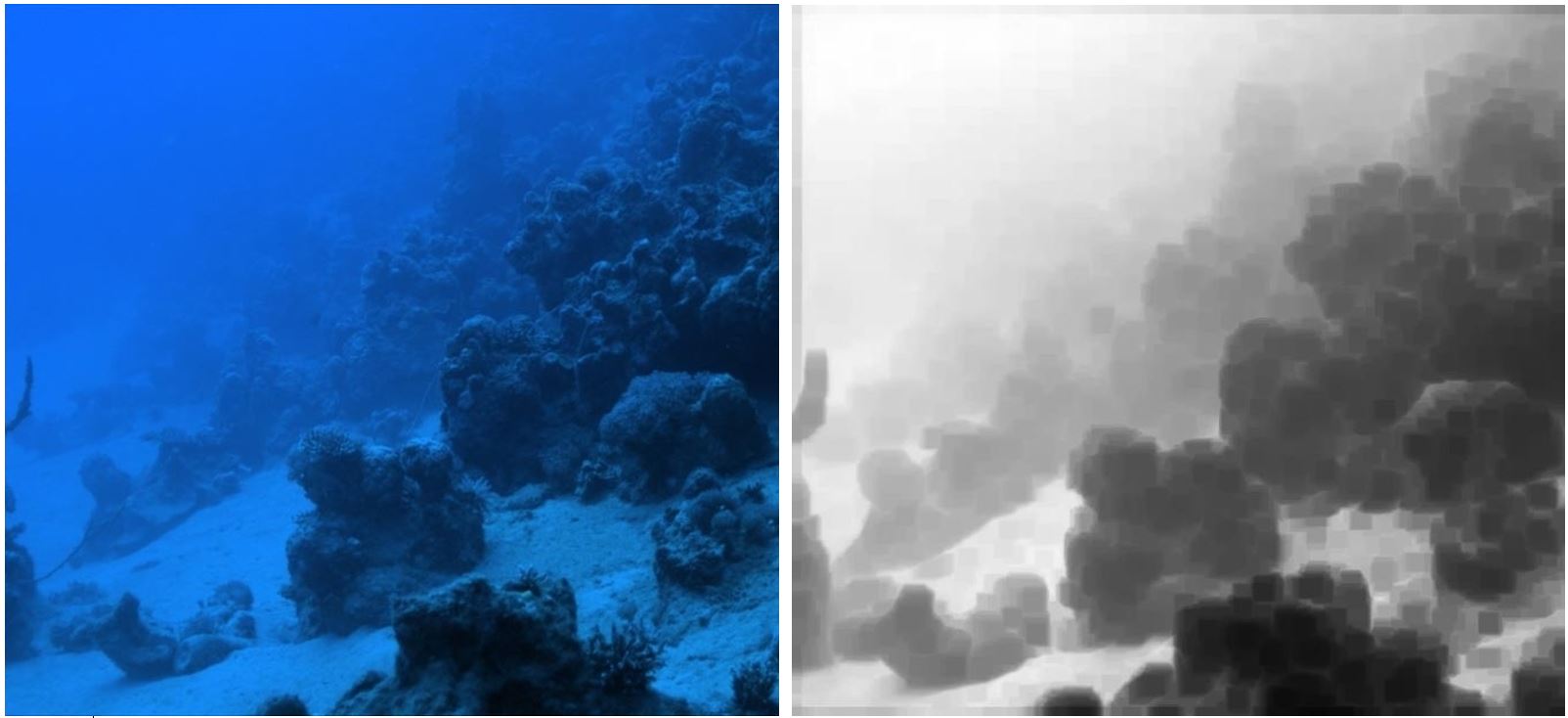}
	\caption{The raw image and the corresponding dark channel calculated with proposed method. Please note that the heavily hazed region in the top left corner of the image is white in the dark channel, correctly indicating very strong presence of the backscattering component.}
	\label{fig:myDC}
\end{figure}

For the image with the shifted RGB coordinate system, its dark channel is calculated. As visible in Fig. \ref{fig:myDC}, the calculated dark channel represents the veiling light in the image much better than the other methods (compare Fig. \ref{fig:UDC} and \ref{fig:RDC}). At this point, $A$ may be estimated by taking the color corresponding to the brightest value in the dark channel. After that, the original image is normalized and processed without any modifications to the original method.

\subsection{First results}
\label{sec:results}
\begin{figure}[htbp]
	\centering
	\includegraphics[width=\linewidth]{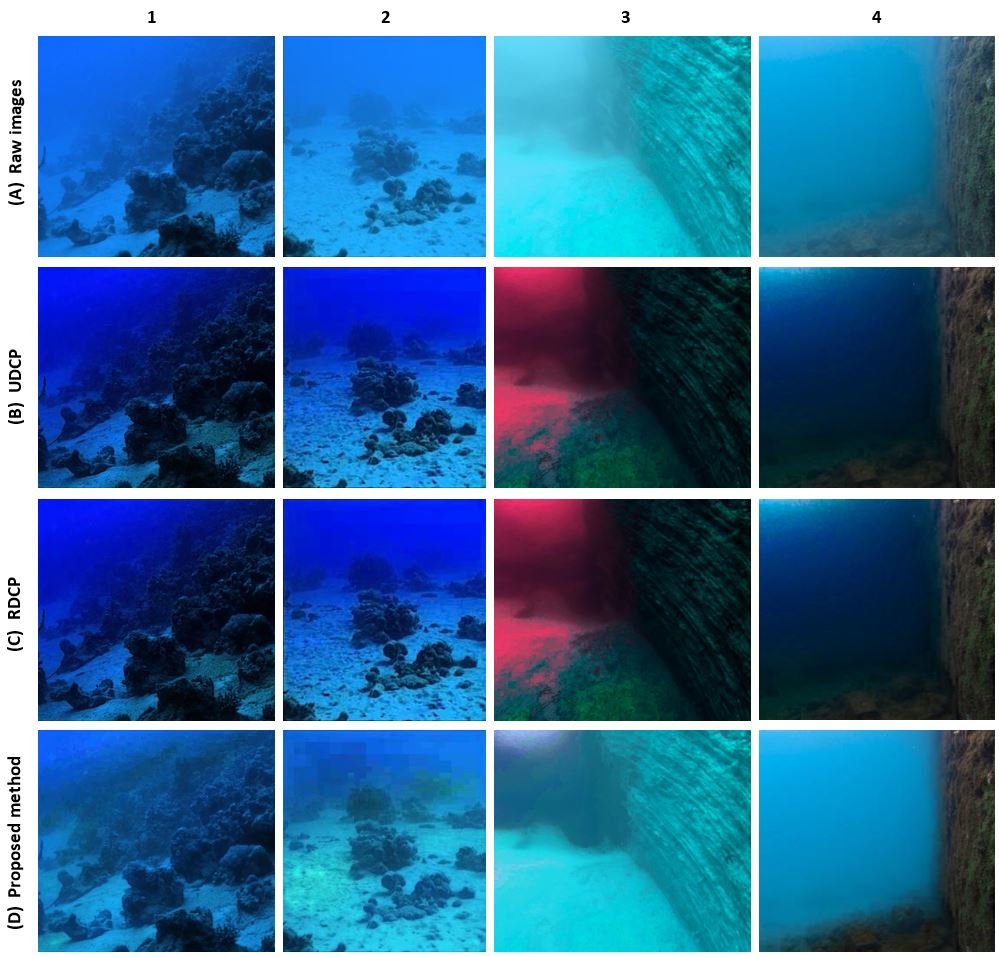}
	\caption{Comparison on different example images. Image 1 and 2 are taken from \cite{SchechnerKarpel2005,SchechnerAverbuch2007}, shared by authors for non-commercial use: http://webee.technion.ac.il/~yoav/research/underwater.html. Images 3 and 4 were recorded during MORPH project trials on the Azores. Following rows present results of image correction with different methods.}
	\label{fig:resultsAll}
\end{figure}

Let's compare the method discussed in the previous section with RDCP and UDCP. A small set of sample images is processed with all three methods for direct comparison. The results are presented on Fig. \ref{fig:resultsAll}.

Depending on the contrast of the raw image, the recovery results may vary. In the worst case, the gain in quality is minimal or none, but the colors after our correction are never distorted, as it happened for RDCP and UDCP, e.g., with sample image 3. Furthermore, there is almost no loss in brightness. Overall the results are promising, as they outperform competing dark channel based methods. The adaptation proposed in this paper adapts the DCP to underwater conditions, preserving all the advantages of this method. On the other hand all the drawbacks of the original work are also present here: computation takes a lot of time and, for some images that do not fulfil statistical assumptions underlying DCP, the results may be faulty.

\section{Final algorithm}
\label{sec:final_alg}
\begin{figure}[htbp]
	\centering
	\includegraphics[width=.9\linewidth]{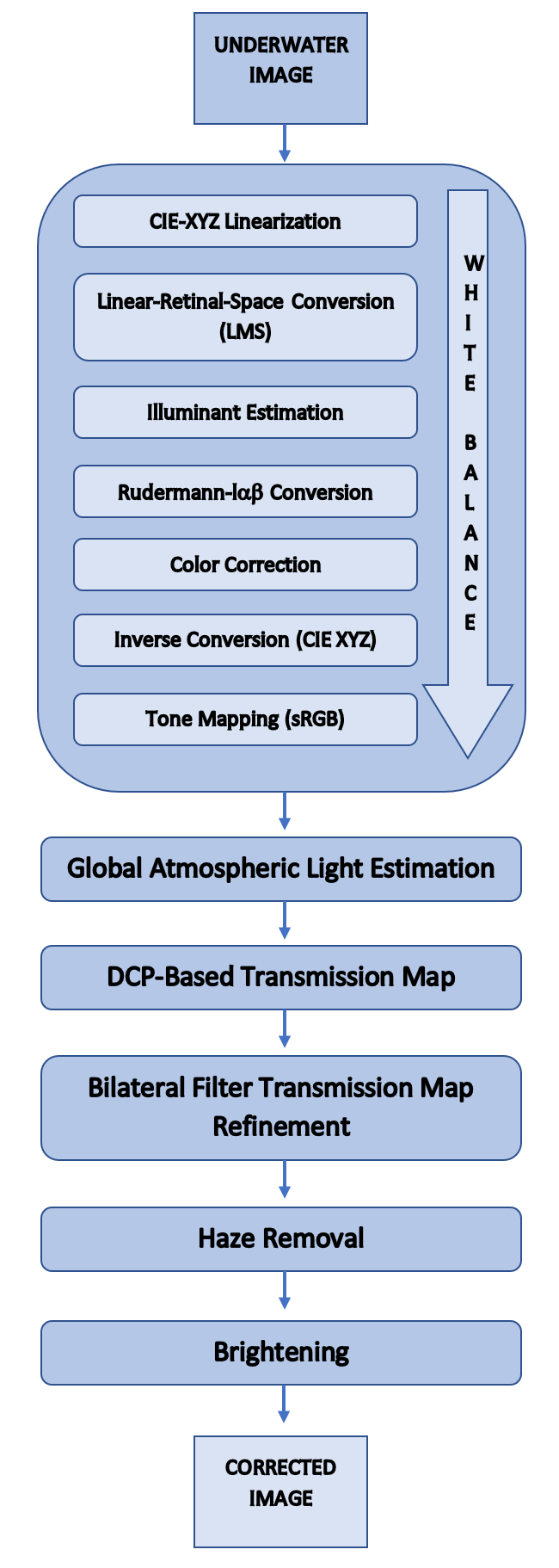}
	\caption{Proposed image processing pipeline. White balance, that can be applied underwater (\cite{BiancoNeumannUW_WB}) allows for global atmospheric light and transmission map estimation like presented in \cite{HeDCP}. Comparing to standard DCP, bilateral filter is being used instead of soft matting to speed up the processing time.}
	\label{fig:pipeline}
\end{figure}

Analysing the initial results two things must be noted. First of all the Dark Channel Prior, as presented here, cannot be used for online processing. This has been discussed before in other papers \cite{bilat_filter_DCP} and the most computationally expensive part od this algorithm is soft matting. This can be improved e.g. by using bilateral filter instead. With this modification DCP can be applied fast enough to be used e.g. for video processing.
The second major issue is that the original assumption regarding the veiling light being white, was substituted by the assumption that underwater it is blue. Even though in many cases it is true, more generalized approach would be desired. A properly applied white balance may be a good solution here. With the grey world assumption it will transform colors in the image so that the veiling light should be white or grey, depending on the intensity of light. However, as discussed in Section \ref{sec:image_formation}, application of the white balance is not straight-forward. Authors or \cite{BiancoNeumannUW_WB} proposed a working solution. White balance is performed in the Rudermann color space, which allows for accurate correction even when color shift is significant. Furthermore, the correction is calculated within a patch surrounding given pixel. Therefore the effect of changing distance is limited and can be only noticed in regions with sharp, step change in distance, e.g. on the edges of some underwater structure with open water behind it (see Fig. \ref{fig:dexrovmockup}). In great majority of cases this not the issue, and even when this happens, it only influences the colours locally and the atmospheric light may still be accurately estimated. 
Therefore after applying white balance the global atmospheric light may be estimated and DCP-based haze removal can be applied. Finally, as the haze correction decreases the overall brightness of the image significantly, the image is brightened. The image processing pipeline proposed here is presented on Fig. \ref{fig:pipeline}.

\section{Results}
\label{sec:final_results}
The algorithm presented here was tested on three video datasets from different locations. All three were selected for a very high level of haze that is present in the images. The correction applied here is not aiming at purely cosmetic changes, but to significantly improve the usability of the data recorded in unfavourable conditions.
Full video comparing registered and improved images from all three datasets may be found on youtube: \textit{youtu.be/f9rcdzBVwC8}

\subsection{An ROV cage}
\begin{figure}[htbp]
	\centering
	\includegraphics[width=\linewidth]{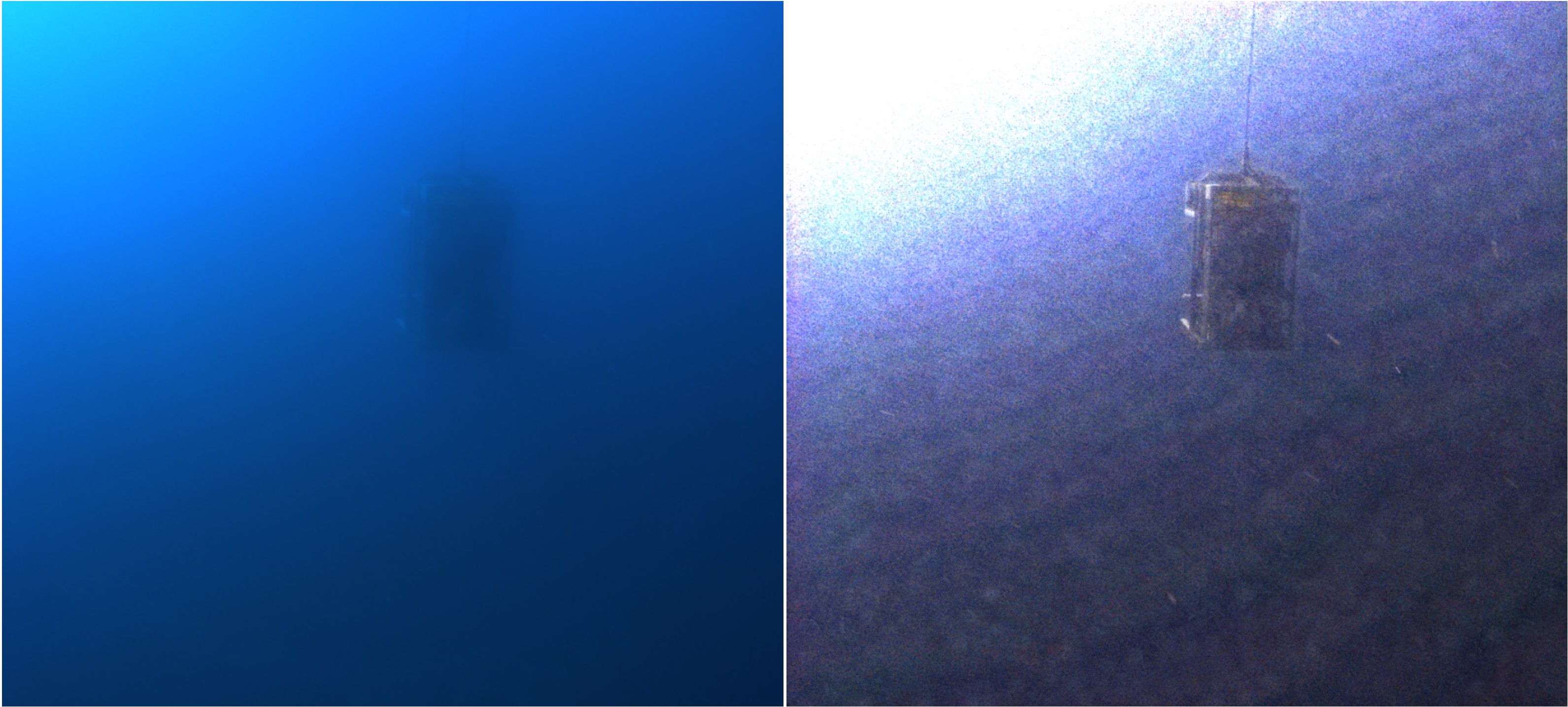}
	\fbox{\includegraphics[width=.95\linewidth]{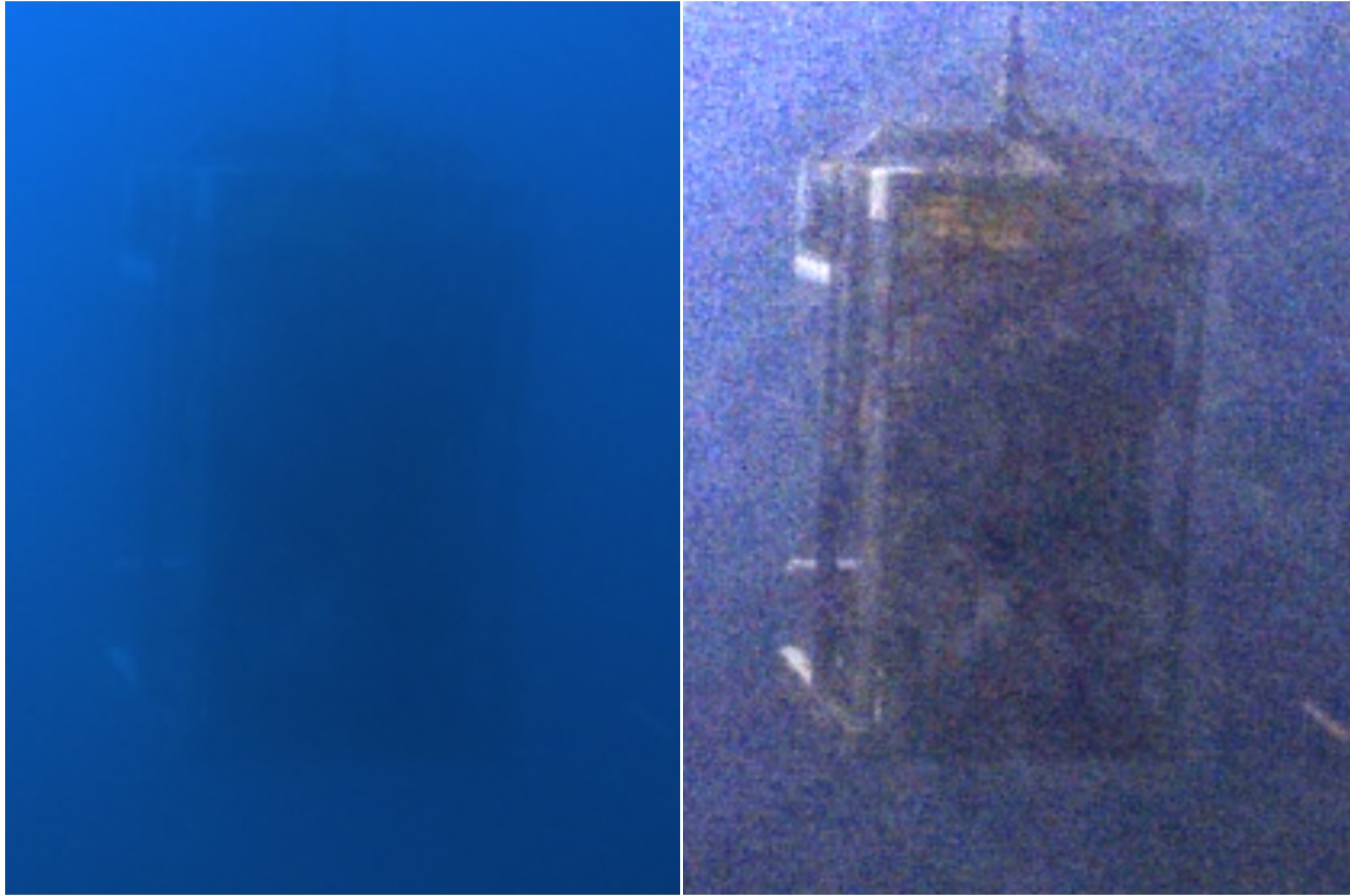}}
	\caption{An ROV cage in the Mediterranean sea. Top: raw and corrected image. Bottom: close up view on the cage in both images.}
	\label{fig:dexrovcage}
\end{figure}

The first dataset is a video of an ROV cage in an open water. It was recorded during the DexROV project trials, in the Mediterranean sea. A vehicle with the camera approaches from the distance, searching for it. In the first frames the cage is barely visible in the raw footage and even when the vehicle gets closer it is hard to recognize any details - only a shape in the water. Figure \ref{fig:dexrovcage} presents raw and corrected frame from the video and a close up view on the ROV cage in these images.

In the corrected footage the cage is not only much easier to find, but also all the details of the structure can be recognized. Some noise and grain shows up in the image, but this does not compromise the significant gain in the amount of information that can be recognized in the corrected image.

\subsection{The Gnalic shipwreck archaeological site}
\begin{figure}[htbp]
	\centering
	\includegraphics[width=\linewidth]{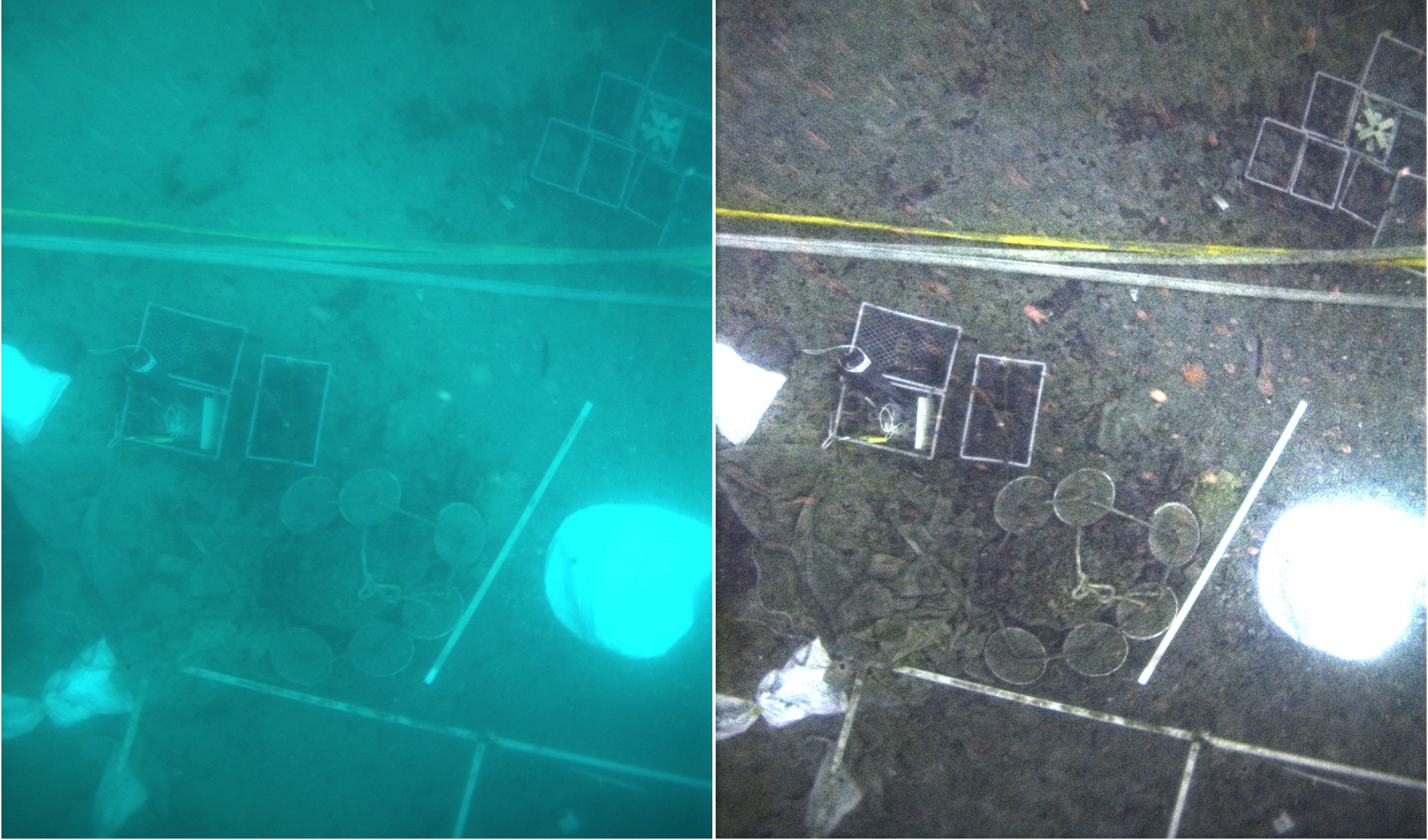}
	\fbox{\includegraphics[width=.95\linewidth]{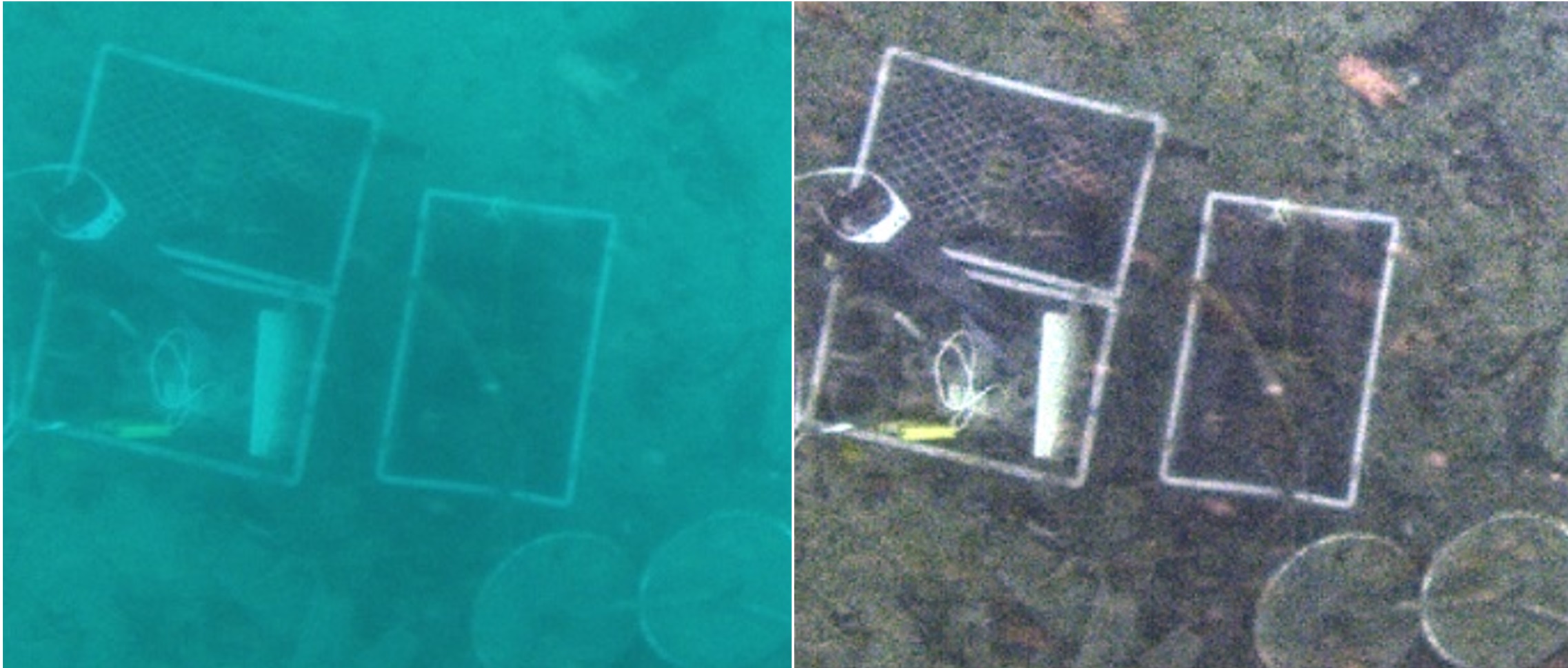}}
	\caption{The Gnalic shipwreck archaeological site in the Adriatic sea. Top: raw and corrected image. Bottom: close up view on the box with tools in both images.}
	\label{fig:gnalic}
\end{figure}

The second dataset comes from the Adriatic sea. The camera was hand held by a diver, slowly descending towards an archaeological site of the Gnalic shipwreck. On that day, due to the strong winds and rough sea, the visibility conditions were poor. In the beginning of the video hardly any details can be recognized. Later, when the camera gets closer, more can be seen. Figure \ref{fig:gnalic} shows a raw frame from the video, the same frame corrected and a close up view on the part of the image with a lot of details. 

In the corrected image more details can be recognised. Please note the mesh on the box with tools - it is barely visible in the raw footage and can be clearly seen in the corrected frame. Since the scene was roughly equidistant to the camera there are also very few artefacts from applying the white balance.

\subsection{The DexROV mockup}
\label{sec:mockup_discussion}
\begin{figure}[htbp]
	\centering
	\includegraphics[width=\linewidth]{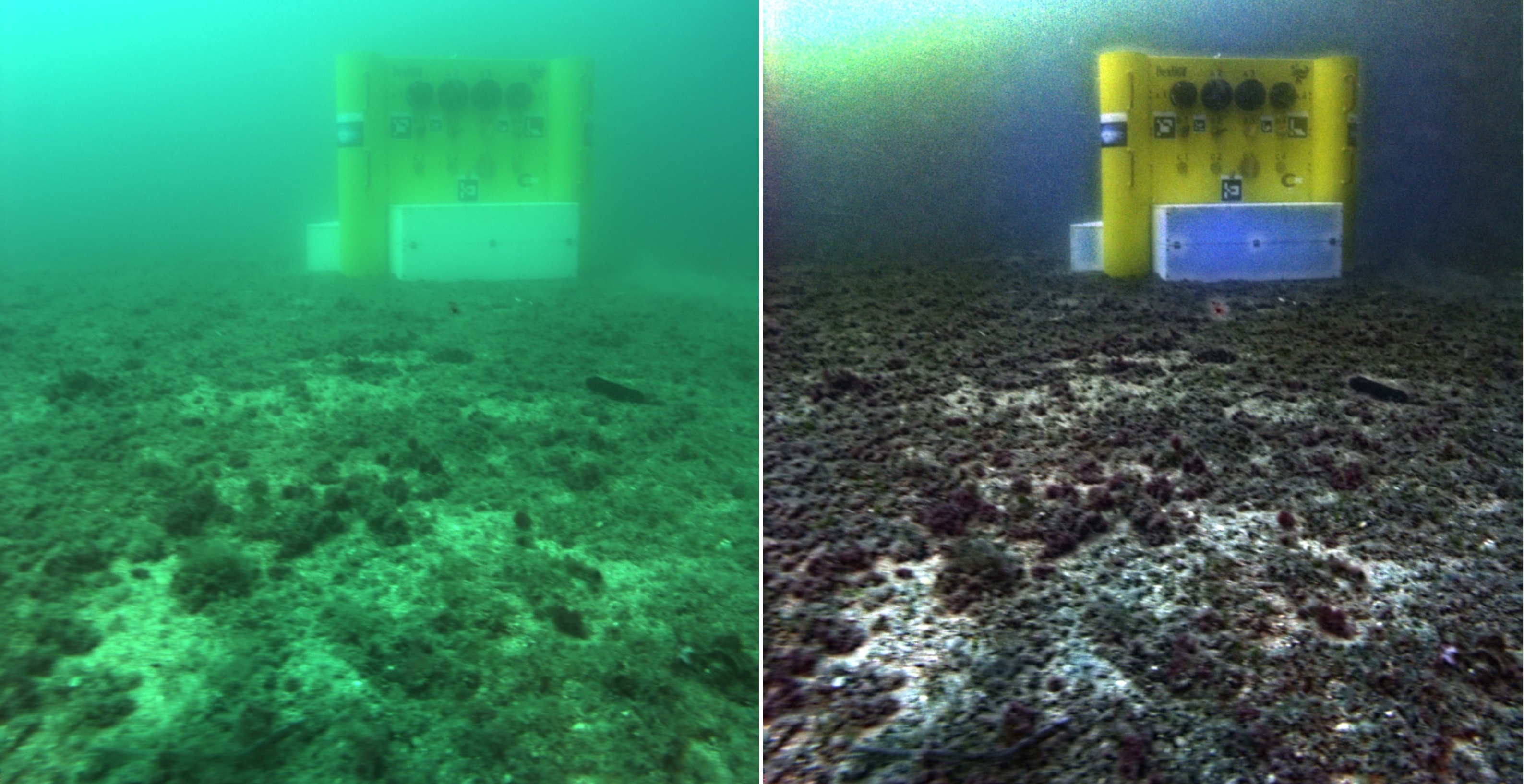}
	\fbox{\includegraphics[width=.95\linewidth]{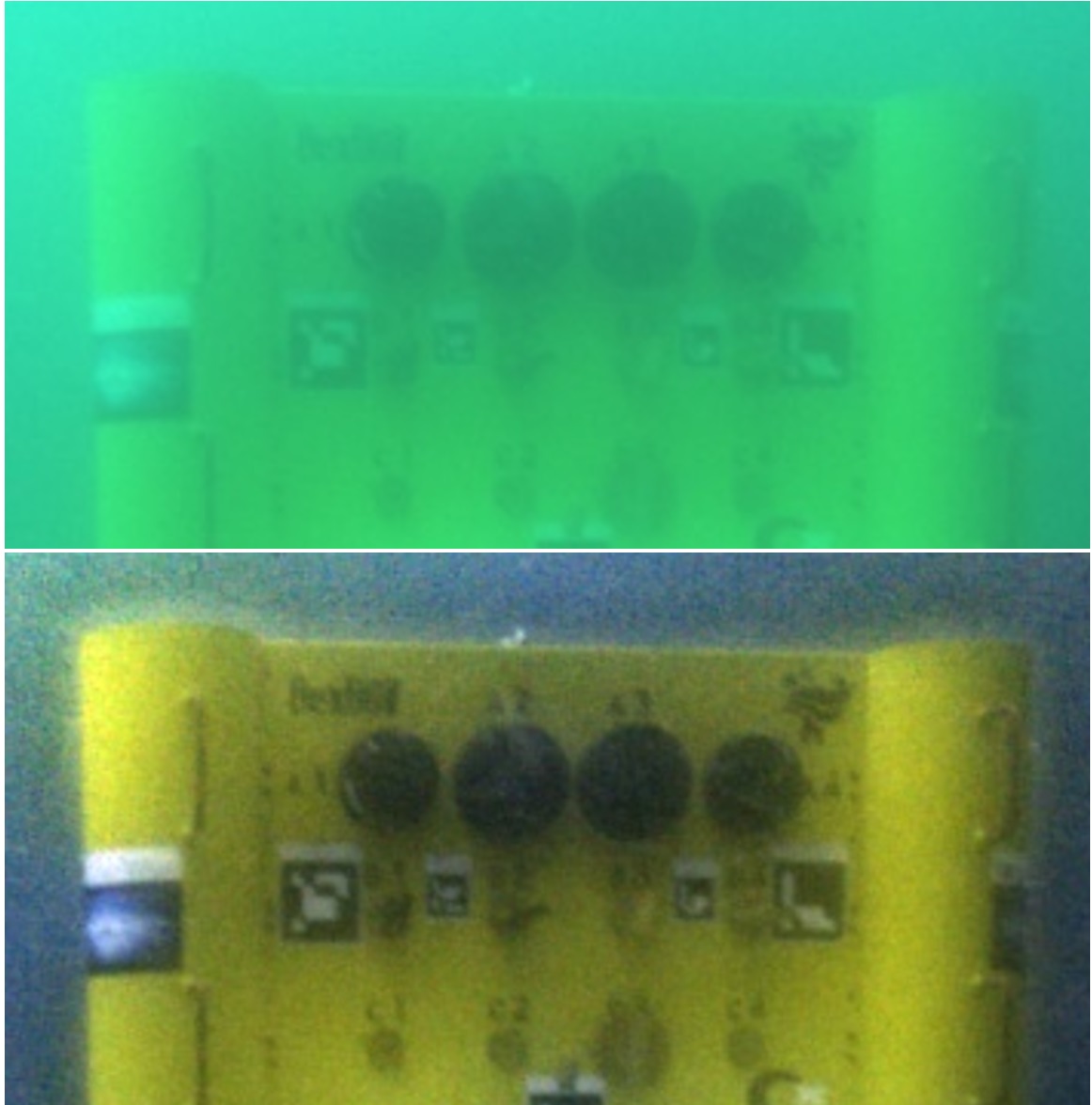}}
	\caption{The DexROV project mockup. Top: raw and corrected image. Bottom: close up view on the test panel in both images.}
	\label{fig:dexrovmockup}
\end{figure}

The third video was also recorded during the DexROV project trials in the Mediterranean sea. The vehicle was moving around the mockup used e.g. for testing manipulation. Figure \ref{fig:dexrovmockup} follows the same scheme as before, showing raw and corrected frames, as well as a close up view below. Please note, that in this case the water is significantly more green than in the previous datasets.
The correction algorithm handles the green hue very well, showing much better robustness than the initial approach presented in \cite{oceansAnchorageDCP} and shortly discussed in Section \ref{sec:results}. Furthermore, many more details can be recognised, e.g. DexROV logo or labels next to the valves and handles on the panel. Some artefacts from white balance are visible in the distorted color, especially on the edges, where the distance to the scene changes rapidly. Furthermore, the distortion of color can be noticed on the white parts of the mockup, where the image was overexposed and any residual noise there was magnified. That being noted, the corrected image offers a significantly more information about the scene, comparing to the raw footage.

\section{Conclusions}
\label{sec:conclusions} 
The Dark Channel Prior (DCP), originally introduced for haze removal in (long-range) outdoor images, is a very useful technique that can be of interest to enhance underwater imagery. Within this paper, a new method of applying the Dark Channel Prior to underwater conditions was presented. 
Our method is based on the insights that (a) the stronger the backscattering component, the more blue/green the region gets and that (b) this property is only important for the correct estimation of the global atmospheric light $A$. Therefore, the core of DCP does not need to be modified as in other state of the art attempts to apply DCP to underwater images, but the method of estimating $A$ needs to be changed instead. 
Tests on real life data from different water bodies were performed. Results show very good performance and robustness against different conditions and water properties. 

\section*{Acknowledgements}
The authors would like to thank prof. Irena Radic Rossi from University of Zadar for her help with collecting data of the Gnalic shipwreck.
The research leading to the presented results was supported in part by the European Community's Horizon2020 program under grant agreement n.~635491 ``Dexterous ROV: effective dexterous ROV operations in presence of communication latencies (DexROV)''.

\bibliographystyle{IEEEtran}
\bibliography{ref}
\end{document}